\documentclass[letterpaper]{article} %
\usepackage[preprint]{aaai2027}  %
\usepackage[hyphens]{url}  %
\usepackage{graphicx} %
\usepackage{natbib}  %
\usepackage{caption} %
\usepackage{algorithm}
\usepackage{algorithmic}
\usepackage{amsmath}
\usepackage{amssymb} 
\usepackage{bm}      
\usepackage{latexsym}
\usepackage{booktabs}
\usepackage[table]{xcolor}
\usepackage{multirow}
\usepackage{array}

\usepackage{subcaption}  
\usepackage{color}
\usepackage{soul}
\usepackage{newfloat}
\usepackage{listings}
\DeclareCaptionStyle{ruled}{labelfont=normalfont,labelsep=colon,strut=off} %
\floatstyle{ruled}
\newfloat{listing}{tb}{lst}{}
\floatname{listing}{Listing}
\title{NaviGATE: Noise-Aware Dual-View Generation for\\
Multimodal Knowledge Graph Reasoning}

\author{
    Jie Peng,
    Yongxue Shan,
    Yongfu Zha,
    Cui Miao,
    Xiaodong Wang
}

\affiliations{
    College of Computer Science and Technology\\
    National University of Defense Technology\\
    Changsha 410073, China\\
    \{pengjie, shanyongxue001, zhayongfu, miaocui1024, xdwang\}@nudt.edu.cn
}

\begin{document}

\maketitle

\begin{abstract}
Multimodal knowledge graph reasoning (MKGR) aims to infer missing facts by jointly exploiting relational structures, textual descriptions, and visual evidence. Existing embedding-based methods effectively capture graph topology, but their fixed multimodal representations are less effective for query-specific fine-grained semantic reasoning, while generation-based methods are vulnerable to noisy retrieval and may produce invalid entities. To address these challenges, we propose NaviGATE, a noise-aware generative framework for MKGR. NaviGATE first aligns graph-derived structural representations with textual semantics, then performs structure-aware internal knowledge activation by fine-tuning the multimodal language model with multimodal entity summaries and injected structural tokens. In the noise-aware dual-view instruction tuning stage, NaviGATE jointly optimizes retrieval-free and retrieval-aware views: reliable retrieved candidates are reinforced through contrastive logit sharpening, while misleading candidates are suppressed through parametric consistency distillation. In inference, structure-constrained decoding restricts generation to valid entities in the target knowledge graph. Experiments on standard benchmarks show that NaviGATE consistently outperforms strong unimodal, multimodal, and generation-based baselines, and remains robust under noisy retrieval scenarios.
\end{abstract}

\section{Introduction}
Multimodal knowledge graphs (MMKGs) integrate structured relational facts with multimodal information, such as textual descriptions and visual images, and have become important knowledge resources for artificial intelligence applications, including recommendation systems \cite{ref3,ref2}, question answering \cite{ref1}, and information retrieval \cite{ref4, ref5}. However, real-world knowledge graphs are often incomplete, with many valid facts missing from the observed graph. Multimodal knowledge graph reasoning (MKGR) aims to predict missing entities by jointly exploiting graph structure and entity-associated multimodal context, and has attracted increasing attention in recent years.

Existing MKGR methods can be broadly categorized into embedding-based and generation-based approaches. Embedding-based methods encode entities, relations, and multimodal attributes into continuous representation spaces, and predict missing facts by scoring candidate triples. With graph neural networks \cite{ref12,ref8}, translation-based constraints \cite{ref9,ref11}, or multimodal fusion modules \cite{ref6,ref7,ref10}, these methods effectively preserve structural topology and enrich entity representations with textual and visual evidence. However, their dependence on fixed multimodal representations and predefined scoring functions limits their ability to perform query-specific fine-grained semantic reasoning. As a result, when multiple candidates are semantically similar, embedding-based models may narrow down the search space but remain insufficient to distinguish the exact target entity.

Generation-based methods \cite{ref13,ref14} offer a promising alternative by formulating MKGR as a text generation problem. Leveraging the semantic understanding and parametric knowledge of multimodal large language models (MLLMs), these methods can generate missing entities from textualized queries and multimodal contexts. Compared with embedding-based models, they provide greater flexibility in contextual reasoning and can better exploit relation semantics, entity descriptions, and visual information. Nevertheless, direct generation may produce answers that are not valid entities in the target knowledge graph or are inconsistent with the graph schema, which limits their completion performance. Recent retrieval-augmented frameworks \cite{ref16,ref15,ref17} mitigate this issue by using an embedding-based retriever to obtain a top-K candidate set, which is then provided to the MLLM for reranking or generation. By combining structural recall with semantic reasoning, these methods have achieved notable performance improvements. However, this pipeline is fundamentally constrained by retriever recall: if the ground-truth entity is absent from the top-K candidate set, MLLMs may fail to reliably generate the correct answer; meanwhile, feeding the entire entity set into the model is impractical due to their limited context windows and attention budgets.

In realistic MKGR scenarios, retrieved candidates are not always reliable, since the ground-truth entity may be absent from the candidate set, making the retrieved candidate set noisy. Over-reliance on retrieval may force the generator to select an incorrect entity from the candidate set. This raises a key challenge: \textbf{how can a generative MKGR model rely on retrieved evidence when it is reliable, while falling back on the LLM's internal knowledge when the retrieved candidate set is noisy?}

To address this challenge, we propose NaviGATE, a progressively optimized generative framework for MKGR. First, structure-semantic collaborative alignment coordinates structural entity representations with textual semantics in a shared space, enabling graph-derived knowledge to be integrated into the MLLM. Second, structure-aware internal knowledge activation trains the model with multimodal entity summaries and structural cues under a retrieval-free view, preserving its basic completion ability when retrieval is noisy. Third, noise-aware dual-view instruction tuning jointly optimizes retrieval-free and retrieval-aware predictions: reliable candidates are strengthened through contrastive logit sharpening, while misleading candidates are suppressed through parametric consistency distillation. During inference, structure-constrained decoding restricts outputs to valid KG entities. The main contributions of this work are summarized as follows:
\begin{itemize}
    \item We propose structure-semantic collaborative alignment and internal knowledge activation to inject graph-derived knowledge into the MLLM, building retrieval-free completion ability as a reliable fallback under noisy retrieval.

    \item We introduce noise-aware dual-view instruction tuning, which strengthens reliable candidates via contrastive logit sharpening and suppresses misleading ones via parametric consistency distillation. Structure-constrained decoding further ensures valid entity generation.

    \item Experiments on MKG-W and DB15K show that NaviGATE achieves the best MRR and Hits@1 among fifteen baselines and remains robust under noisy retrieval.
\end{itemize}

\section{Related Work}
\subsection{Embedding-based knowledge graph reasoning}
Embedding-based knowledge graph reasoning (KGR) learns entity and relation representations and predicts missing facts by scoring candidate triples. Classical models, such as TransE~\cite{ref18}, DistMult~\cite{ref19}, ComplEx~\cite{ref20}, RotatE~\cite{ref21}, and TuckER~\cite{ref22}, mainly exploit structural triples to model relational patterns. MKGR methods further incorporate images and textual descriptions into entity representation learning, including IKRL~\cite{ref23}, RSME~\cite{ref24}, OTKGE~\cite{ref25}, IMF~\cite{ref12}, QEB~\cite{ref27}, VISTA~\cite{ref28}, NativE~\cite{ref6}, MyGO~\cite{ref7}, and APKGR~\cite{ref9}. These methods improve structural retrieval through modality fusion, interaction, adaptive learning, or noise modeling, but still rely on fixed representations and predefined scoring functions. As fine-grained multimodal clues are compressed before inference, their query-specific reasoning ability remains limited when candidates are semantically similar.

\subsection{Generation-based knowledge graph reasoning}
Generation-based KGR formulates link prediction as a text generation task, where a generative language model produces the missing entity from a textualized query. Early methods, such as GenKGR~\cite{ref30} and KG-S2S~\cite{ref13}, convert triples into natural language sequences and apply sequence-to-sequence models for entity generation. With the emergence of large language models, recent studies further explore instruction tuning, in-context learning, and retrieval-augmented generation for KGR. For instance, KG-LLM~\cite{ref29} and KoPA~\cite{ref14} finetunes LLMs on KG completion instructions, while KICGPT~\cite{ref15}, SAT~\cite{ref17}, and DIFT~\cite{ref16} combine embedding-based retrieval with LLM-based reranking or finetuning. These methods demonstrate the potential of language models in leveraging semantic priors for KGR. Nevertheless, they often remain sensitive to retrieval quality. When retrieved candidates are noisy or the ground-truth entity is absent, generation models can be misled by irrelevant evidence. Moreover, directly generated answers may not always correspond to valid entities in the KG. In contrast, our framework targets multimodal KGR by combining noise-aware dual-view instruction tuning with structure-constrained inference, achieving robust and valid entity generation under noisy multimodal retrieval.

\begin{figure*}[t]
\centering
\includegraphics[width=0.9\linewidth]{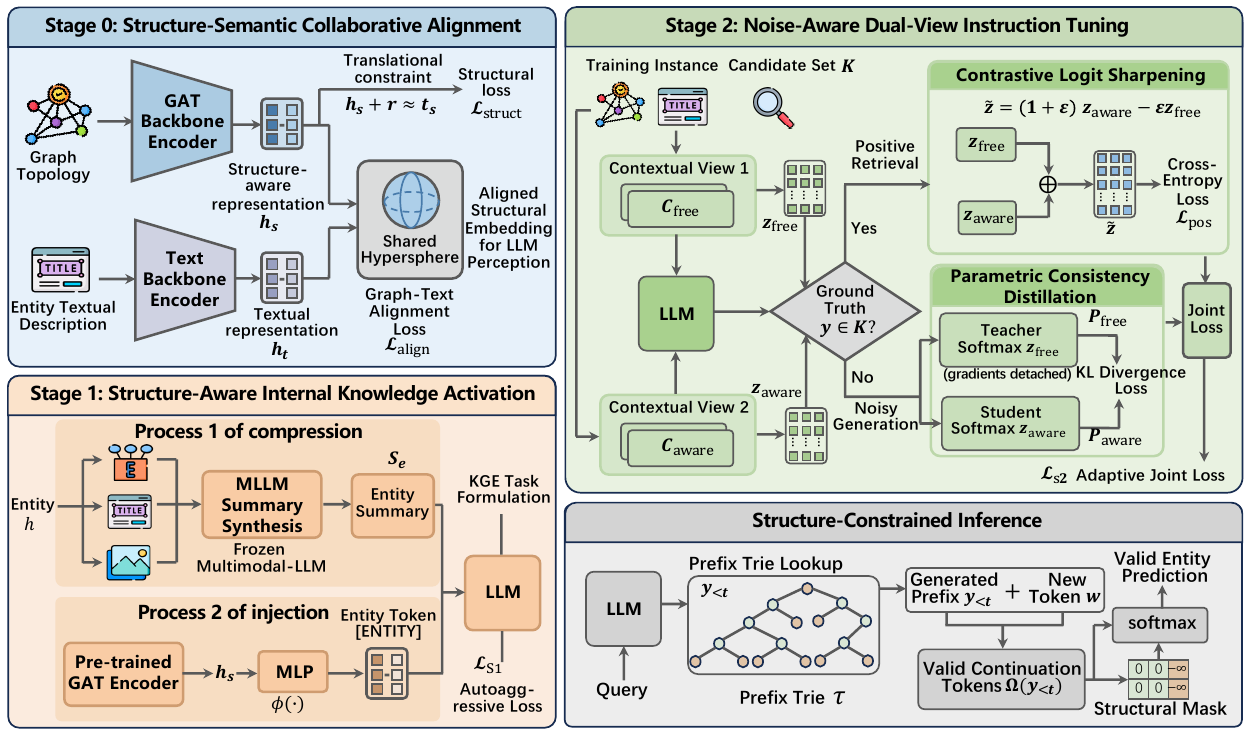}
\caption{Overview of NaviGATE. The framework consists of three progressive training stages: structure-semantic collaborative alignment, structure-aware internal knowledge activation, and noise-aware dual-view instruction tuning. At inference time, structure-constrained decoding ensures valid entity generation.}
\label{fig:NaviGATE_framework}
\end{figure*}

\section{Methodology}
\label{sec:method}
As shown in Figure~\ref{fig:NaviGATE_framework}, NaviGATE progressively performs structure-semantic collaborative alignment, retrieval-free internal knowledge activation, and noise-aware dual-view instruction tuning to exploit reliable candidates while mitigating misleading ones. Fin ally, structure-constrained decoding ensures valid entity generation.

\subsection{Problem Formulation}
\label{sec:problem}

We denote a multimodal knowledge graph as
\begin{equation}
\mathcal{G}
=
\left(
\mathcal{E},
\mathcal{R},
\mathcal{T},
\mathcal{D},
\mathcal{I}
\right),
\end{equation}
where $\mathcal{E}$ and $\mathcal{R}$ denote the entity set and relation set, respectively.
$\mathcal{T}=\{(h,r,t)\mid h,t\in\mathcal{E}, r\in\mathcal{R}\}$ is the set of observed triples.
Each entity $e\in\mathcal{E}$ may be associated with a textual description $d_e\in\mathcal{D}$ and a visual image $i_e\in\mathcal{I}$. Given a query triple with one missing entity, MKGR aims to infer the missing entity from the entity set. Depending on whether the missing entity occurs in the tail or head position, the completion query can be formulated as $(h,r,\ast)$ or $(\ast,r,t)$, respectively, where $\ast$ denotes the unknown entity.

Unlike conventional embedding-based MKGR methods that score candidate triples with fixed representation functions, NaviGATE formulates MKGR as a constrained generation problem. Given the query context $C_q$, the model directly generates the surface name of the missing entity:
\begin{equation}
P_{\theta}(y\mid C_q)
=
\prod_{\ell=1}^{|Y_y|}
P_{\theta}
\left(
Y_{y,\ell}
\mid
Y_{y,<\ell}, C_q
\right),
\end{equation}
where $Y_y=(Y_{y,1},\ldots,Y_{y,|Y_y|})$ denotes the token sequence of entity $y$ under the tokenizer vocabulary $\mathcal{W}$. During inference, the generation space is constrained to valid entity names in $\mathcal{E}$.

\subsection{Structure-Semantic Collaborative Alignment}
\label{sec:alignment}

A central difficulty in generative MKGR is that graph topology and language semantics are represented in fundamentally different spaces. To bridge this gap, NaviGATE first learns structure-aware entity representations and aligns them with textual semantic representations.

\paragraph{Structure-aware entity encoding.}
We use a graph encoder $f_g(\cdot)$ to encode each entity into a structure-aware representation. For entity $e_i$, its representation is obtained by aggregating information from its local neighborhood $\mathcal{N}_i$:
\begin{equation}
\mathbf{s}_i
=
\sigma\bigl(
\sum_{j\in\mathcal{N}_i\cup\{i\}}
\alpha_{ij}
\mathbf{W}_g
\mathbf{x}_j
\bigr),
\end{equation}
where $\mathbf{x}_j$ is the initial feature of entity $e_j$, $\mathbf{W}_g$ is a learnable transformation matrix, $\alpha_{ij}$ is the attention weight between $e_i$ and $e_j$, and $\sigma(\cdot)$ is a activation function.

To preserve relational geometry, we further impose a translation-based structural objective \cite{ref31}. For each positive triple $(h,r,t)\in\mathcal{T}$, we construct a corrupted negative triple $(\bar{h},r,\bar{t})$ by replacing either the head or the tail entity. The structural loss is defined as
\begin{equation}
\mathcal{L}_{\mathrm{kg}}
=
\sum_{(h,r,t)\in\mathcal{T}}
\left[
\gamma
+
d(\mathbf{s}_h+\mathbf{r},\mathbf{s}_t)
-
d(\mathbf{s}_{\bar{h}}+\mathbf{r},\mathbf{s}_{\bar{t}})
\right]_+ .
\end{equation}
where $\mathbf{r}$ denotes the relation embedding, $d(\cdot,\cdot)$ is implemented as the $L_1$ distance, $\gamma$ is the margin, and $[x]_+=\max(0,x)$. This objective encourages valid triples to have lower energy than corrupted triples, thereby preserving graph topology in the structural embedding space.

\paragraph{Contrastive structure–text alignment.}
Although $\mathbf{s}_i$ captures graph topology, it resides in a 
geometry-driven space that is misaligned with the semantic space of 
language models. To bridge this gap, we ground the structural 
representation of each entity in its textual semantics. Concretely, 
we encode the entity description $d_i$ with a frozen language encoder to obtain the textual representation $\mathbf{u}_i$, and map both representations into a shared space:
\begin{equation}
\bar{\mathbf{s}}_i
=
\mathrm{Norm}
\left(
g_s(\mathbf{s}_i)
\right),
\quad
\bar{\mathbf{u}}_i
=
\mathrm{Norm}
\left(
g_t(\mathbf{u}_i)
\right),
\end{equation}
where $g_s(\cdot)$ and $g_t(\cdot)$ are learnable projection heads, and $\mathrm{Norm}(\cdot)$ denotes $L_2$ normalization.

We adopt a symmetric contrastive objective to align matched graph-text pairs while separating mismatched pairs within the same mini-batch. Let $\mathcal{B}$ denote a mini-batch, and define the InfoNCE loss as
\begin{equation}
\ell_{\mathrm{NCE}}(\mathbf{a}_i,\mathbf{b}_i)
=
-
\log
\frac{
\exp
\left(
\mathrm{sim}(\mathbf{a}_i,\mathbf{b}_i)/\tau
\right)
}{
\sum_{j\in\mathcal{B}}
\exp
\left(
\mathrm{sim}(\mathbf{a}_i,\mathbf{b}_j)/\tau
\right)
},
\end{equation}
where $\mathrm{sim}(\cdot,\cdot)$ denotes cosine similarity and $\tau$ is a temperature parameter. The symmetric graph-text contrastive loss is written in expectation form as
\begin{equation}
\mathcal{L}_{\mathrm{con}}
=
\frac{1}{2}
\mathbb{E}_{i\sim\mathcal{B}}
\left[
\ell_{\mathrm{NCE}}
\left(
\bar{\mathbf{s}}_i,\bar{\mathbf{u}}_i
\right)
+
\ell_{\mathrm{NCE}}
\left(
\bar{\mathbf{u}}_i,\bar{\mathbf{s}}_i
\right)
\right].
\end{equation}

The overall alignment objective is
\begin{equation}
\mathcal{L}_{\mathrm{align}}
=
\mathcal{L}_{\mathrm{kg}}
+
\lambda_{\mathrm{con}}
\mathcal{L}_{\mathrm{con}},
\end{equation}
where $\lambda_{\mathrm{con}}$ balances the contributions of the structural and contrastive objectives.

\subsection{Structure-Aware Internal Knowledge Activation}
\label{sec:activation}

Retrieval-augmented methods depend heavily on the quality of the retrieved candidate set. However, in realistic MKGR scenarios, the ground-truth entity may be absent due to limited retriever recall. To reduce such over-dependence on retrieval, NaviGATE first activates the model's retrieval-free completion ability before exposing it to retrieved candidates.

\paragraph{Query-aware multimodal entity summarization.}
Raw entity descriptions, images, and neighboring triples can be lengthy, redundant, and noisy. Directly concatenating them into the prompt may cause context dilution and make it difficult for the MLLM to identify query-relevant evidence. Therefore, for each entity $e$ involved in query $q$, we construct a compact query-aware multimodal summary:
\begin{equation}
S_e^q
=
\mathrm{Summarize}
\left(
q,
d_e,
i_e,
\mathcal{N}_e
\right),
\end{equation}
where $d_e$ is the textual description, $i_e$ is the associated image, and $\mathcal{N}_e$ denotes the local structural neighborhood of $e$. The summary $S_e$ converts heterogeneous evidence into a concise textual form, highlighting entity attributes, visual cues, and graph-neighborhood semantics relevant to query.

\paragraph{Structural token injection.}
To explicitly provide the MLLM with graph structural information, we project the aligned structural representation into the input embedding space of the MLLM. Specifically, for each entity $e$, we compute
\begin{equation}
\mathbf{p}_e
=
\phi(\mathbf{s}_e),
\end{equation}
where $\phi(\cdot)$ is a lightweight multilayer perceptron that matches the dimension of $\mathbf{s}_e$ to that of the MLLM token embeddings. The projected vector $\mathbf{p}_e$ is treated as a soft structural token and inserted into the input embedding sequence.

For a tail prediction query $q=(h,r,\ast)$ as an example, the retrieval-free context is constructed as
\begin{equation}
C_q^{\mathrm{free}}
=
\mathrm{Prompt}
\left(
q,
S_h^q,
\mathbf{p}_h
\right),
\end{equation}
where $S_h^q$ provides query-aware multimodal evidence for the observed entity, while $\mathbf{p}_h$ encodes its structural context. The context excludes retrieved candidate entities, encouraging the model to predict the missing entity based on its internal parametric knowledge and the provided multimodal and structural evidence.

\paragraph{Retrieval-free warm-up objective.}
The model is optimized to generate the ground-truth entity name from the retrieval-free context:
\begin{equation}
\mathcal{L}_{\mathrm{act}}
=
-
\sum_{\ell=1}^{|Y_y|}
\log
P_{\theta}
\left(
Y_{y,\ell}
\mid
Y_{y,<\ell},
C_q^{\mathrm{free}}
\right).
\end{equation}
This stage equips the model with retrieval-independent completion ability by learning to predict the missing entity from its parametric knowledge and the provided multimodal-structural evidence. It also establishes a retrieval-free prediction distribution that later serves as a stable reference for suppressing misleading candidate evidence.

\subsection{Noise-Aware Dual-View Instruction Tuning}
\label{sec:dual_view}

Building on the retrieval-free completion ability established during internal knowledge activation, NaviGATE is further optimized with retrieved candidates. Let $\mathcal{K}_q={k_1,\ldots,k_K}$ denote the top-K candidate entities retrieved for query $q$ by an embedding-based retriever. These candidates may provide useful non-parametric evidence, but the ground-truth entity may be absent and the remaining candidates may be misleading. Therefore, the model should learn to exploit retrieved evidence when it is reliable, while preserving its internal parametric knowledge when retrieval is noisy.

To this end, we construct two complementary views for each training query:
\begin{equation}
\begin{aligned}
C_q^{\mathrm{free}}
&=
\operatorname{Prompt}(q,S_h,\mathbf{p}_h),\\
C_q^{\mathrm{aware}}
&=
\operatorname{Prompt}(q,S_h,\mathbf{p}_h,\mathcal{K}_q).
\end{aligned}
\end{equation}
The retrieval-free view captures the model's internal belief without external candidates, while the retrieval-aware view incorporates retrieved entity evidence.

Let $Y_y=(Y_{y,1},\ldots,Y_{y,|Y_y|})$ denote the token sequence of the ground-truth entity.
At each decoding position $\ell$, under teacher forcing with prefix $Y_{y,<\ell}$, the model produces vocabulary logits for each view:
\begin{equation}
\mathbf{z}_{\theta,\ell}^{v}
=
\operatorname{LMHead}
\left(
\mathbf{h}_{\theta,\ell}^{v}
\right)
\in
\mathbb{R}^{|\mathcal{V}|},
\quad
v\in\{\mathrm{free},\mathrm{aware}\},
\end{equation}
where $\mathbf{h}_{\theta,\ell}^{v}$ is the hidden state conditioned on context $C_q^v$ and prefix $Y_{y,<\ell}$, and $\mathcal{V}$ denotes the vocabulary.

\paragraph{Sharpened generative supervision for reliable retrieval.}\mbox{}\\
When the ground-truth entity is included in the retrieved set, i.e., $y\in\mathcal{K}_q$, the retrieved candidates provide useful external evidence. In this case, the retrieval-aware logits should be encouraged to amplify the preference gain induced by the retrieved context.

Specifically, for each decoding position $\ell$, we first compute the retrieval-induced logit shift:
\begin{equation}
\Delta \mathbf{z}_{\theta,\ell}
=
\mathbf{z}_{\theta,\ell}^{\mathrm{aware}}
-
\operatorname{sg}
\left[
\mathbf{z}_{\theta,\ell}^{\mathrm{free}}
\right],
\end{equation}
where $\operatorname{sg}[\cdot]$ denotes the stop-gradient operation.
We then construct sharpened retrieval-aware logits:
\begin{equation}
\widetilde{\mathbf{z}}_{\theta,\ell}
=
\mathbf{z}_{\theta,\ell}^{\mathrm{aware}}
+
\epsilon
\Delta \mathbf{z}_{\theta,\ell},
\end{equation}
where $\epsilon\geq0$ controls the sharpening strength.
Equivalently,
\begin{equation}
\widetilde{\mathbf{z}}_{\theta,\ell}
=
(1+\epsilon)
\mathbf{z}_{\theta,\ell}^{\mathrm{aware}}
-
\epsilon
\operatorname{sg}
\left[
\mathbf{z}_{\theta,\ell}^{\mathrm{free}}
\right].
\end{equation}
This operation amplifies the logit-level preference shift from the retrieval-free view to the retrieval-aware view, thereby strengthening the useful signal introduced by reliable retrieval.

The reliable-retrieval loss is computed as the token-level negative log-likelihood under the sharpened logits:
\begin{equation}
\mathcal{L}_{\mathrm{rel}}
=
-
\frac{1}{|Y_y|}
\sum_{\ell=1}^{|Y_y|}
\log
\frac{
\exp
\left(
\widetilde{z}_{\theta,\ell,Y_{y,\ell}}
\right)
}{
\sum_{w\in\mathcal{V}}
\exp
\left(
\widetilde{z}_{\theta,\ell,w}
\right)
}.
\end{equation}
This objective encourages the model to enlarge the relative advantage of the ground-truth entity under the retrieval-aware view, while suppressing candidate preferences that mainly originate from the retrieval-free prior.

\paragraph{Parametric consistency distillation for noisy retrieval.}\mbox{}\
When the ground-truth entity is absent from the retrieved set, i.e., $y\notin\mathcal{K}_q$, applying logit sharpening to the retrieval-aware prediction may bias the model toward plausible but incorrect candidates. Therefore, we use the retrieval-free branch as a parametric anchor and constrain the retrieval-aware branch at each decoding position.

For each output position $\ell$, we minimize the KL divergence between the retrieval-free token distribution and the retrieval-aware token distribution:
\begin{equation}
\begin{aligned}
\mathcal{L}_{\mathrm{pcd}}
&=
\frac{1}{|Y_y|}
\sum_{\ell=1}^{|Y_y|}
D_{\mathrm{KL}}
\left(
\operatorname{sg}
\left[
\mathbf{p}_{\theta,\ell}^{\mathrm{free}}
\right]
\,\middle\|\,
\mathbf{p}_{\theta,\ell}^{\mathrm{aware}}
\right) \\
&=
\frac{1}{|Y_y|}
\sum_{\ell=1}^{|Y_y|}
\sum_{w\in\mathcal{V}}
\operatorname{sg}
\left[
p_{\theta,\ell,w}^{\mathrm{free}}
\right]
\log
\frac{
\operatorname{sg}
\left[
p_{\theta,\ell,w}^{\mathrm{free}}
\right]
}{
p_{\theta,\ell,w}^{\mathrm{aware}}
},
\end{aligned}
\end{equation}
where
\begin{equation}
\mathbf{p}_{\theta,\ell}^{v}
=
\operatorname{softmax}
\left(
\mathbf{z}_{\theta,\ell}^{v}
\right),
\quad
v\in\{\mathrm{free},\mathrm{aware}\}.
\end{equation}
The stop-gradient operation prevents the consistency objective from updating the retrieval-free branch, while encouraging the retrieval-aware branch to resist misleading retrieval evidence and remain consistent with the model's internal parametric knowledge.
Finally, the dual-view tuning objective is
\begin{equation}
\mathcal{L}_{\mathrm{dual}}
=
\mathbb{I}
\left[
y\in\mathcal{K}_q
\right]
\mathcal{L}_{\mathrm{rel}}
+
\lambda_{kl}
\mathbb{I}
\left[
y\notin\mathcal{K}_q
\right]
\mathcal{L}_{\mathrm{pcd}},
\end{equation}
where $\lambda_{kl}$ balances the parametric consistency objective.

\subsection{Structure-Constrained Inference}
\label{sec:inference}

Unconstrained generation may produce entity names that do not exist in the
target knowledge graph. Since MKGR requires each prediction to correspond to
a valid entity in $\mathcal{E}$, we introduce a structure-constrained decoding
strategy to ensure entity validity.

We first construct a prefix trie $\mathcal{P}$ from the tokenized names of all
entities in $\mathcal{E}$, with an end-of-sequence token appended to each
entity name. At decoding step $t$, given the generated prefix $Y_{<t}$, the
set of valid next tokens is defined as
\begin{equation}
\Omega(Y_{<t})
=
\left\{
w \in \mathcal{V}
\;\middle|\;
Y_{<t} \circ w
\text{ is a valid prefix in } \mathcal{P}
\right\},
\end{equation}
where $\circ$ denotes token concatenation. We then define the vocabulary
constraint mask as
\begin{equation}
M_t(w)
=
\begin{cases}
0, & w \in \Omega(Y_{<t}), \\
-\infty, & w \notin \Omega(Y_{<t}).
\end{cases}
\end{equation}

During inference, NaviGATE computes the retrieval-free and retrieval-aware
token logits, denoted by $\mathbf{z}_{\theta,t}^{\mathrm{free}}$ and
$\mathbf{z}_{\theta,t}^{\mathrm{aware}}$, respectively. Both predictions are
conditioned on the generated prefix $Y_{<t}$. We define the retrieval-induced
logit shift as
\begin{equation}
\Delta \mathbf{z}_{\theta,t}
=
\mathbf{z}_{\theta,t}^{\mathrm{aware}}
-
\mathbf{z}_{\theta,t}^{\mathrm{free}}.
\end{equation}

Following the logit-level adjustment used during training, the inference
logits are computed as
\begin{equation}
\widetilde{\mathbf{z}}_{\theta,t}
=
\mathbf{z}_{\theta,t}^{\mathrm{aware}}
+
\eta \Delta \mathbf{z}_{\theta,t},
\end{equation}
where $\eta \geq 0$ controls the adjustment strength. The final decoding
distribution is defined as
\begin{equation}
P_{\theta}
\left(
Y_t = w
\mid
Y_{<t}, q
\right)
=
\frac{
\exp\left(
\widetilde{z}_{\theta,t,w} + M_t(w)
\right)
}{
\displaystyle
\sum_{v \in \mathcal{V}}
\exp\left(
\widetilde{z}_{\theta,t,v} + M_t(v)
\right)
}.
\end{equation}

The prefix constraint ensures that each completed output corresponds to a
valid entity name in the target KG. When retrieved candidates provide useful
evidence, the logit adjustment strengthens the retrieval-induced preference.
Under noisy retrieval, parametric consistency distillation reduces the
discrepancy between the retrieval-free and retrieval-aware predictions,
thereby limiting the influence of misleading candidates.

\begin{table*}[t]
\centering
\caption{Overall results on MKG-W and DB15K. The best results are shown in \textbf{bold}, and the second-best results are \underline{underlined}. We reproduce the results of RotatE, TuckER, DIFT, and SAT using their released code. Results for the remaining baselines are taken from their respective original papers.}
\setlength{\tabcolsep}{8pt}
\label{tab:results}
\small
\begin{tabular}{lcccccccc}
\toprule
\multirow{2}{*}{\textbf{Models}} & \multicolumn{4}{c}{\textbf{MKG-W}} & \multicolumn{4}{c}{\textbf{DB15K}} \\
\cmidrule(lr){2-5} \cmidrule(lr){6-9}
& \textbf{MRR} & \textbf{Hits@1} & \textbf{Hits@3} & \textbf{Hits@10} & \textbf{MRR} & \textbf{Hits@1} & \textbf{Hits@3} & \textbf{Hits@10} \\
\midrule
\multicolumn{9}{c}{\textbf{Unimodal KGR Methods}} \\
\midrule
TransE~\citep{ref18} & 0.292 & 0.211 & 0.332 & 0.442 & 0.249 & 0.128 & 0.315 & 0.471 \\
DistMult~\citep{ref19} & 0.210 & 0.159 & 0.223 & 0.309 & 0.230 & 0.148 & 0.263 & 0.396 \\
ComplEx~\citep{ref20} & 0.249 & 0.191 & 0.267 & 0.367 & 0.275 & 0.184 & 0.316 & 0.454 \\
RotatE~\citep{ref21} & 0.334 & 0.270 & 0.362 & 0.451 & 0.291 & 0.184 & 0.355 & 0.479 \\
TuckER~\citep{ref22} & 0.301 & 0.243 & 0.323 & 0.408 & 0.340 & 0.258 & 0.374 & 0.501 \\
\midrule
\multicolumn{9}{c}{\textbf{Discriminative MKGR methods \& Generative Models}} \\
\midrule
OTKGE~\citep{ref25} & 0.344 & 0.289 & 0.363 & 0.449 & 0.239 & 0.185 & 0.259 & 0.342 \\
IMF~\citep{ref12} & 0.345 & 0.288 & 0.366 & 0.454 & 0.323 & 0.242 & 0.360 & 0.482 \\
QEB~\citep{ref27} & 0.324 & 0.255 & 0.351 & 0.453 & 0.282 & 0.148 & 0.367 & 0.516 \\
AdaMF~\citep{ref32} & 0.343 & 0.272 & 0.379 & 0.472 & 0.325 & 0.213 & 0.397 & 0.517 \\
NativE~\citep{ref6} & 0.366 & 0.296 & 0.395 & 0.494 & 0.362 & 0.273 & 0.411 & \textbf{0.529} \\
SnAg~\citep{ref33} & 0.359 & 0.284 & 0.394 & 0.492 & 0.342 & 0.250 & 0.391 & 0.515 \\
MyGO~\citep{ref7} & 0.361 & 0.300 & 0.383 & 0.469 & \underline{0.372} & 0.296 & \underline{0.413} & 0.520 \\
APKGR~\citep{ref9} & 0.374 & 0.306 & 0.404 & \underline{0.501} & 0.364 & 0.282 & \underline{0.413} & \underline{0.527} \\
DHNS~\citep{ref35} & 0.346 & 0.268 & 0.392 & \textbf{0.512} & 0.364 & 0.297 & 0.398 & 0.492 \\
DIFT+TuckER~\citep{ref16}  & 0.361 & 0.317 & 0.390 & 0.442 & 0.367 & 0.304 & 0.402 & 0.505 \\
DIFT+RotatE~\citep{ref16}  & \underline{0.390} & \underline{0.351} & \underline{0.416} & 0.468 & 0.332 & 0.267 & 0.366 & 0.482 \\
SAT+TuckER~\citep{ref17} & 0.363 & 0.320 & 0.388 & 0.441 & 0.369 & \underline{0.305} & 0.401 & 0.503 \\
SAT+RotatE~\citep{ref17} & 0.389 & 0.347 & 0.415 & 0.471 & 0.336 & 0.271 & 0.369 & 0.485 \\
\midrule
\textbf{NaviGATE+TuckER}  & 0.378 & 0.330 & 0.403 & 0.453 & \textbf{0.383} & \textbf{0.313} & \textbf{0.415} & 0.516 \\
\textbf{NaviGATE+RotatE}  & \textbf{0.408} & \textbf{0.361} & \textbf{0.430} & 0.487 & 0.348 & 0.279 & 0.379 & 0.491 \\
\bottomrule
\end{tabular}
\end{table*}

\section{Experiments}
\subsection{Experimental Setup}
\paragraph{Datasets.}
We evaluate NaviGATE on two widely used multimodal knowledge graph reasoning benchmarks, namely DB15K and MKG-W. DB15K contains 12,842 entities, 279 relations, 79,222 training triples, 9,902 validation triples, and 9,904 test triples. MKG-W contains 15,000 entities, 169 relations, 34,196 training triples, 4,276 validation triples, and 4,274 test triples. Both datasets provide structural triples together with multimodal entity information, making them suitable for evaluating the ability of MKGR models.

\paragraph{Baselines.}
We compare NaviGATE with three categories of baseline methods. First, we include representative unimodal knowledge graph embedding models that rely only on structural triples, including TransE~\cite{ref18}, DistMult~\cite{ref19}, ComplEx~\cite{ref20}, RotatE~\cite{ref21}, and TuckER~\cite{ref22}. Second, we compare with discriminative MKGR methods that incorporate visual or textual information into entity representation learning, including OTKGE~\cite{ref25}, IMF~\cite{ref12}, QEB~\cite{ref27}, AdaMF~\cite{ref32}, SnAg~\cite{ref33}, NativE~\cite{ref6}, MyGO~\cite{ref7}, and APKGR~\cite{ref9}. Third, we include generation-based KGR methods, including DIFT~\cite{ref16} and SAT~\cite{ref17}, to evaluate whether NaviGATE can leverage generative reasoning while maintaining robustness to noisy retrieval results.

\paragraph{Implementation Details.}
We conduct all experiments on an Intel Xeon Gold CPU with 4 NVIDIA 4090 GPUs, each with 24 GB memory, under Ubuntu 18.04 LTS. We select TuckER and RotatE as representative embedding-based retrievers. Each retriever is first pretrained on the training set, and the top-20 predicted entities are then retrieved for each query as candidate entities. For the multimodal generative backbone, we use Qwen3-VL 8B as the base model and adopt LoRA~\cite{ref34} for parameter-efficient finetuning. The LoRA rank, scaling factor, and dropout rate are set to 64, 16, and 0.1, respectively. We train the model for 2 epochs in the internal knowledge activation stage and 4 epochs in the dual-view instruction tuning stage. During inference, we use prefix-trie constrained decoding to generate a valid entity, place it at the top of the retriever's original ranking, remove any duplicate occurrence, and preserve the relative order of all remaining entities. We then report standard filtered MRR and Hits@1/3/10.

\paragraph{Main Results.}
Table~\ref{tab:results} reports the overall results. Compared with unimodal KGR methods, NaviGATE consistently achieves better performance on both datasets, demonstrating the effectiveness of integrating multimodal information with generative reasoning for MKGR.

Compared with discriminative MKGR baselines, NaviGATE achieves stronger and more consistent performance on the key top-ranking metrics across both datasets. On MKG-W, NaviGATE+RotatE improves the best MRR and Hits@1 from 0.374 and 0.306 to 0.408 and 0.361, respectively, demonstrating a clear advantage in ranking the correct entity at the top. On DB15K, NaviGATE+TuckER reaches an MRR of 0.383 and a Hits@1 of 0.313, also surpassing the strongest discriminative baselines.

 Compared with LLM-based methods such as DIFT and SAT, NaviGATE achieves further improvements under both retrievers. On MKG-W, NaviGATE+RotatE improves MRR by 1.8 percentage points over DIFT+RotatE, while NaviGATE+TuckER outperforms SAT+TuckER by 1.5 percentage points. On DB15K, NaviGATE+TuckER further improves MRR by 1.4 percentage points over SAT+TuckER. These results show that noise-aware tuning effectively exploits reliable candidates while mitigating noisy retrieval.

\begin{figure}[t]
  \centering
\includegraphics[width=0.90\linewidth]{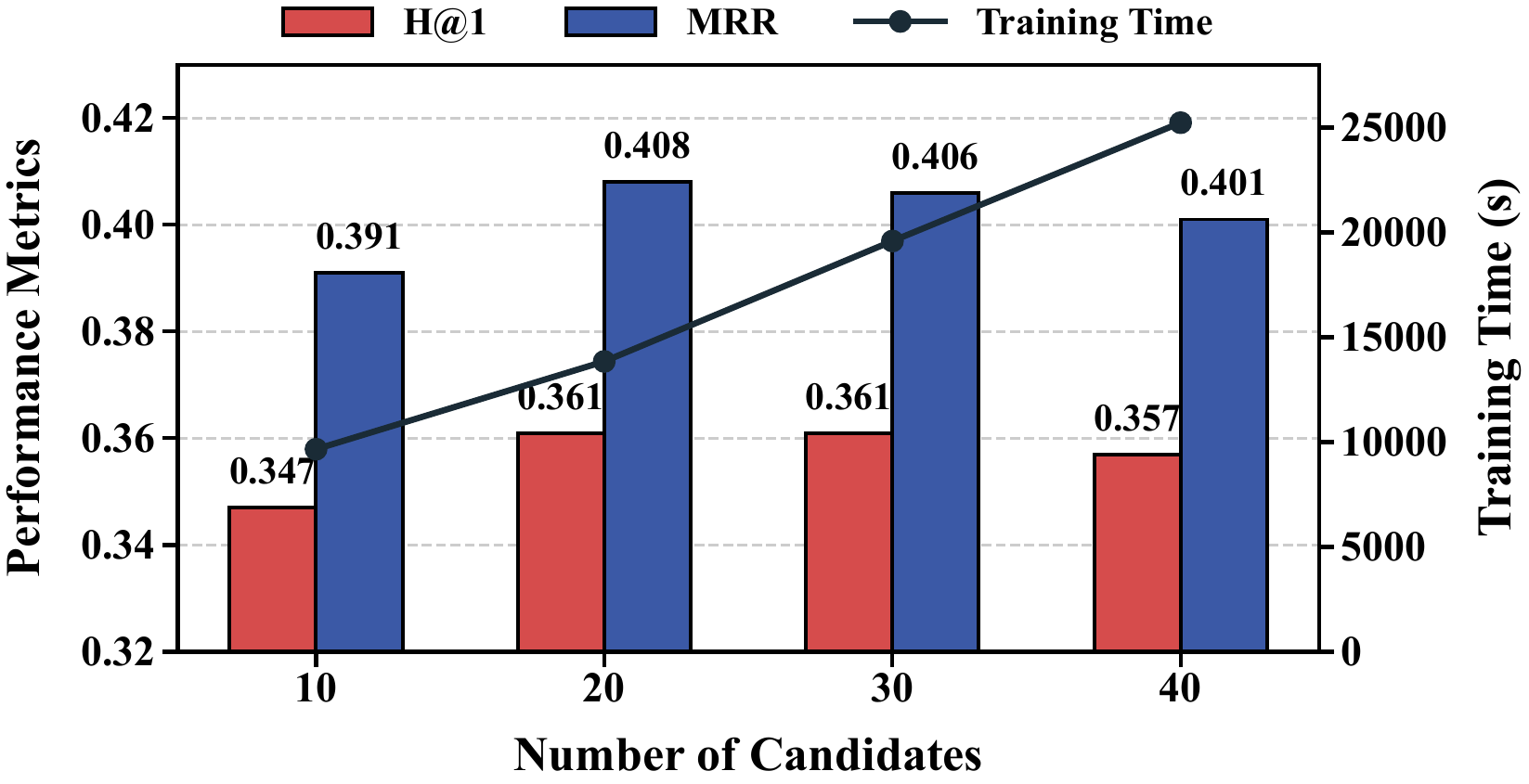}
  \caption{Impact of the number of candidates in prompt inputs on model performance and training time on the MKG-W dataset.}
  \label{fig:candidates_performance}
\end{figure}

\begin{table}[t]
\centering
\caption{Ablation results on MKG-W across training stages, dual-view tuning components, and modality configurations. ``w/o'' denotes removing the corresponding component from the full model. Best results are shown in bold.
}
\label{tab:ablation}
\footnotesize
\setlength{\tabcolsep}{3pt}
\renewcommand{\arraystretch}{1.08}
\begin{tabular}{@{}>{\raggedright\arraybackslash}p{3.8cm}cccc@{}}
\toprule
\textbf{Variants} & \textbf{MRR} & \textbf{H@1} & \textbf{H@3} & \textbf{H@10} \\
\midrule
\multicolumn{5}{@{}l}{\textit{Stage-1 Structural Design}} \\
w/o stage1 pretrain          & 0.401 & 0.357 & 0.421 & 0.478 \\
w/o structural contrastive loss & 0.405 & 0.357 & 0.430 & 0.486 \\
\midrule
\multicolumn{5}{@{}l}{\textit{Stage-2 Reranking Design}} \\
w/o stage2 finetune          & 0.338 & 0.256 & 0.390 & 0.468 \\
w/o multimodal text summary  & 0.387 & 0.334 & 0.411 & 0.478 \\
w/o structure-const. inference & 0.395 & 0.348 & 0.415 & 0.481 \\
w/o KL loss                  & 0.397 & 0.352 & 0.416 & 0.477 \\
w/o retrieval-free           & 0.398 & 0.352 & 0.416 & 0.480 \\
w/o retrieval-aware          & 0.338 & 0.248 & 0.398 & 0.478 \\
\midrule
\multicolumn{5}{@{}l}{\textit{Modality Ablation}} \\
w/o visual modality          & 0.399 & 0.350 & 0.423 & 0.481 \\
w/o textual modality         & 0.364 & 0.307 & 0.389 & 0.466 \\
w/o structural modality      & 0.401 & 0.350 & 0.427 & 0.485 \\ 
\midrule
\multicolumn{5}{l}{\textit{Noisy-Retrieval Subset} ($y \notin K_q$)} \\
w/o parametric consistency & 0.396 & 0.350 & 0.423 & 0.482 \\
\midrule
\textbf{Full Model} & \textbf{0.408} & \textbf{0.361} & \textbf{0.430} & \textbf{0.487} \\ 
\bottomrule
\end{tabular}
\end{table}

\begin{figure}[t]
  \centering
  \includegraphics[width=\linewidth]{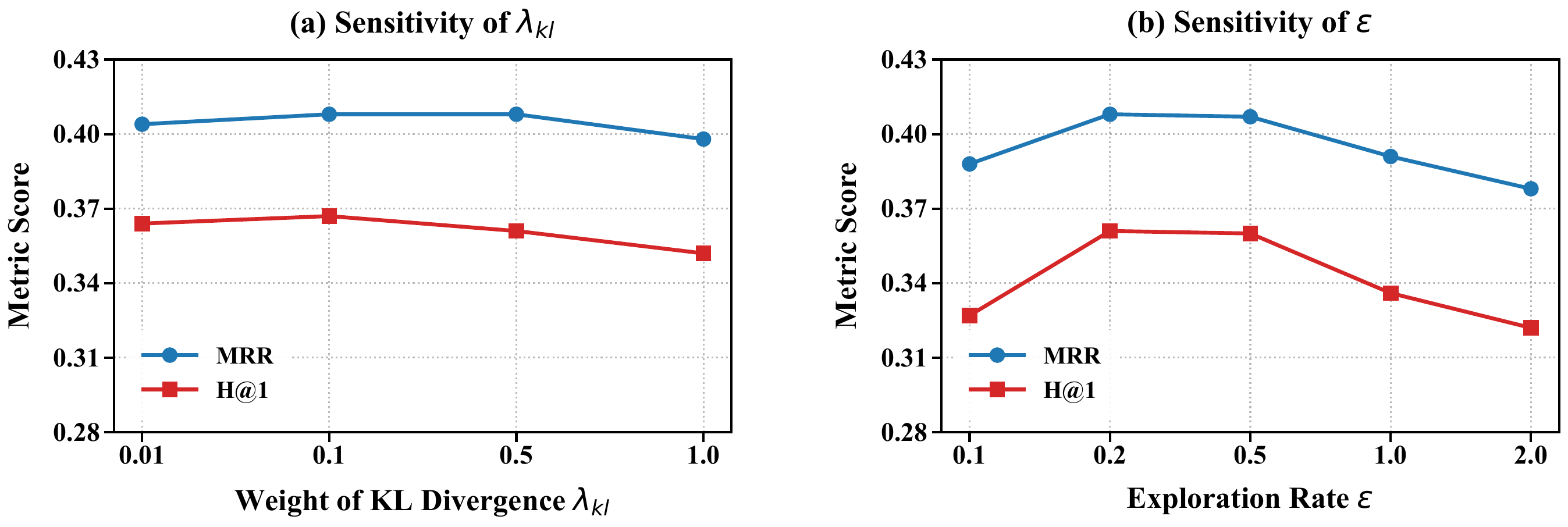}
  \caption{Hyperparameter sensitivity analysis on the MKG-W dataset: (a) Impact of the KL divergence weight $\lambda_{kl}$ under noisy retrieval conditions, and (b) Impact of the exploration rate $\varepsilon$ in contrastive logit sharpening.}
  \label{fig:hyperparameter_sensitivity}
\end{figure}

\paragraph{Ablation Study.}
Table~\ref{tab:ablation} presents the ablation results on MKG-W. Removing stage-1 pretraining or the structural contrastive loss decreases MRR from 0.408 to 0.401 and 0.405, respectively, confirming that structure-semantic alignment provides a useful foundation for subsequent generative reasoning. In the dual-view tuning stage, removing stage-2 finetuning causes a substantial drop in both MRR and Hits@1, while removing the retrieval-aware branch results in the lowest Hits@1 of 0.248. These results indicate that retrieved evidence remains essential when properly incorporated through noise-aware optimization. The multimodal summary, structure-constrained decoding, KL loss, and retrieval-free branch also yield consistent improvements, demonstrating their complementary roles in semantic enrichment, valid entity generation, and robustness to noisy candidates. Among the three modalities, removing textual information leads to the largest degradation, whereas visual and structural information provide smaller but stable gains. To directly evaluate robustness under retrieval failure, we further report results on the subset where the ground-truth entity is absent from the candidate set. As shown in Table~\ref{tab:ablation}, NaviGATE improves Hits@1 by 1.1 percentage points over the variant without parametric consistency, confirming its ability to suppress misleading retrieval evidence.

\begin{figure}[htbp]
  \centering
  \includegraphics[width=\linewidth]{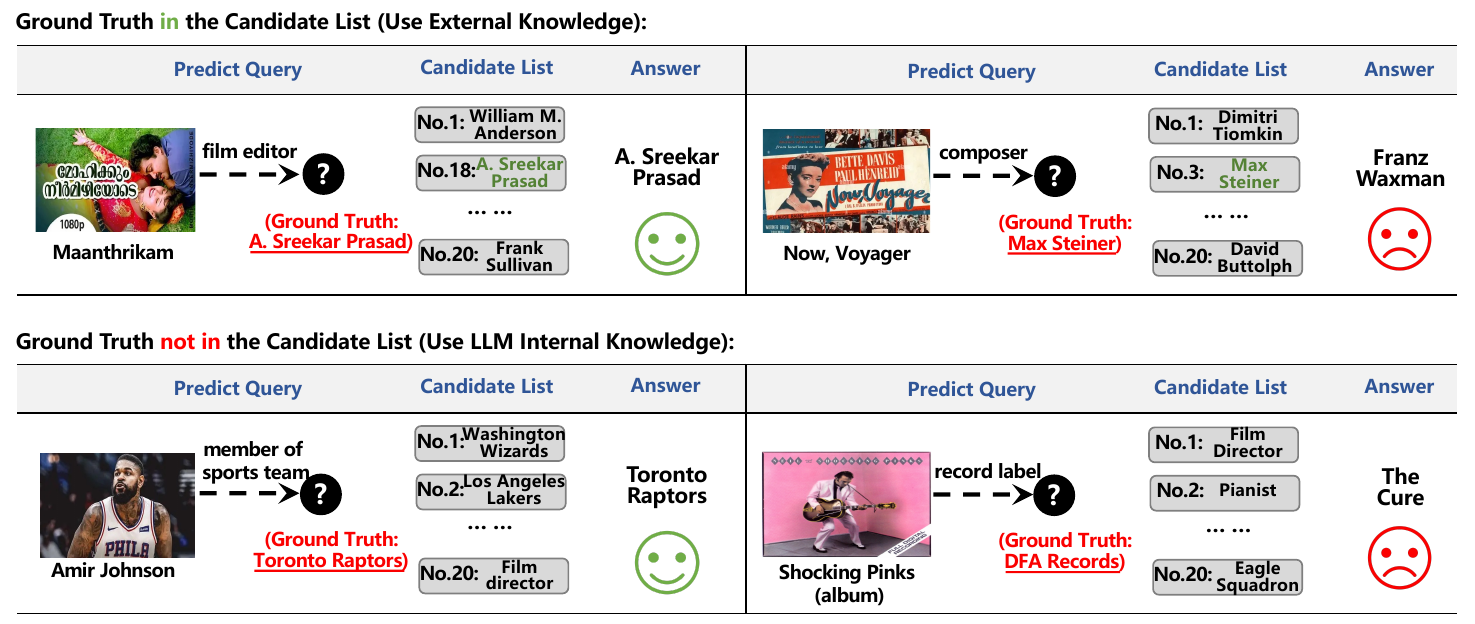}
  \caption{Case study of NaviGATE under different retrieval conditions. The upper part shows examples where the ground-truth entity is included in the candidate list, while the lower part shows examples where it is absent.}
  \label{fig:case_study}
\end{figure}

\paragraph{Impact of the Number of Retrieved Candidates.}
Figure~\ref{fig:candidates_performance} analyzes the effect of candidate number on MKG-W. Increasing candidates from 10 to 20 improves MRR from 0.391 to 0.408 and Hits@1 from 0.347 to 0.361, showing that a moderate candidate set provides more useful retrieval evidence. However, further increasing the number to 30 or 40 brings no additional gain and slightly degrades performance, likely due to more noisy distractors in the prompt. Since training time also increases steadily with more candidates, we use 20 candidates as the default setting for better accuracy-efficiency trade-off.

\paragraph{Hyperparameter Sensitivity.}
Figure~\ref{fig:hyperparameter_sensitivity} shows the sensitivity of NaviGATE to $\lambda_{kl}$ and $\varepsilon$ on MKG-W. For $\lambda_{kl}$, the model performs stably when it ranges from 0.01 to 0.5, while a larger value of 1.0 slightly degrades both MRR and Hits@1. This suggests that moderate parametric consistency helps suppress noisy retrieval, whereas overly strong regularization may limit the benefit of retrieval-aware evidence. For $\varepsilon$, the best performance is achieved around 0.2--0.5. A small $\varepsilon$ provides insufficient logit sharpening, while a large value over-amplifies retrieval-induced shifts and introduces noise. Therefore, we set $\lambda_{kl}=0.1$ and $\varepsilon=0.2$ by default.

\paragraph{Case Study.}
Figure~\ref{fig:case_study} shows representative cases under different retrieval conditions. When the ground-truth entity appears in the candidate list, NaviGATE exploits retrieval evidence to generate the correct answer. More importantly, even when the ground-truth entity is absent from the candidates, NaviGATE can still recover it using activated internal knowledge. The failure cases further show that distinguishing highly confusable entities under ambiguous multimodal cues remains challenging, highlighting the inherent difficulty of noise-aware reasoning in MKGR.

\section{Conclusion}
In this paper, we propose NaviGATE, a noise-aware generative framework for multimodal knowledge graph reasoning. NaviGATE aligns structural representations with textual semantics, activates retrieval-free completion through multimodal summaries and structural token injection, and employs dual-view instruction tuning to exploit reliable candidates while suppressing misleading ones via parametric consistency. Structure-constrained decoding further ensures valid entity generation. Experiments demonstrate its effectiveness and robustness. In future work, we will extend NaviGATE to incorporate more modality information beyond text and images, and explore robust designs for missing-modality scenarios to improve real-world MKGR reliability.

\bibliography{aaai2027}

@inproceedings{ref1,
  title={Boosting visual question answering with context-aware knowledge aggregation},
  author={Li, Guohao and Wang, Xin and Zhu, Wenwu},
  booktitle={Proceedings of the 28th ACM International Conference on Multimedia},
  pages={1227--1235},
  year={2020}
}

@inproceedings{ref2,
  author       = {Yejin Kim and
                  Scott Rome and
                  Kevin Foley and
                  Mayur Nankani and
                  Rimon Melamed and
                  Javier Morales and
                  Abhay Yadav and
                  Maria Peifer and
                  Sardar Hamidian and
                  H. Howie Huang},
  title        = {Improving Content Recommendation: Knowledge Graph-Based Semantic Contrastive
                  Learning for Diversity and Cold-Start Users},
  booktitle    = {Proceedings of the 2024 Joint International Conference on Computational
                  Linguistics, Language Resources and Evaluation},
  pages        = {8743--8755},
  year         = {2024},
}

@inproceedings{ref3,
  author       = {Yikun Xian and
                  Zuohui Fu and
                  S. Muthukrishnan and
                  Gerard de Melo and
                  Yongfeng Zhang},
  title        = {Reinforcement Knowledge Graph Reasoning for Explainable Recommendation},
  booktitle    = {Proceedings of the 42nd International {ACM} {SIGIR} Conference on
                  Research and Development in Information Retrieval},
  pages        = {285--294},
  year         = {2019},
}

@inproceedings{ref4,
  author       = {Zihao Li and
                  Yuyi Ao and
                  Jingrui He},
  title        = {SpherE: Expressive and Interpretable Knowledge Graph Embedding for
                  Set Retrieval},
  booktitle    = {Proceedings of the 47th International {ACM} {SIGIR} Conference on
                  Research and Development in Information Retrieval},
  pages        = {2629--2634},
  year         = {2024},
}

@inproceedings{ref5,
author = {Kim, Juyeon and Lee, Geon and Kim, Taeuk and Shin, Kijung},
title = {KGMEL: Knowledge Graph-Enhanced Multimodal Entity Linking},
year = {2025},
isbn = {9798400715921},
publisher = {Association for Computing Machinery},
address = {New York, NY, USA},
url = {https://doi.org/10.1145/3726302.3730217},
doi = {10.1145/3726302.3730217},
booktitle = {Proceedings of the 48th International ACM SIGIR Conference on Research and Development in Information Retrieval},
pages = {3015–3019},
numpages = {5},
location = {Padua, Italy},
series = {SIGIR '25}
}

@inproceedings{ref6,
  author       = {Yichi Zhang and
                  Zhuo Chen and
                  Lingbing Guo and
                  Yajing Xu and
                  Binbin Hu and
                  Ziqi Liu and
                  Wen Zhang and
                  Huajun Chen},
  title        = {NativE: Multi-modal Knowledge Graph Completion in the Wild},
  booktitle    = {{SIGIR}},
  pages        = {91--101},
  publisher    = {{ACM}},
  year         = {2024}
}

@inproceedings{ref7,
  author       = {Yichi Zhang and
                  Zhuo Chen and
                  Lingbing Guo and
                  Yajing Xu and
                  Binbin Hu and
                  Ziqi Liu and
                  Wen Zhang and
                  Huajun Chen},
  title        = {Tokenization, Fusion, and Augmentation: Towards Fine-grained Multi-modal
                  Entity Representation},
  booktitle    = {{AAAI}},
  pages        = {13322--13330},
  publisher    = {{AAAI} Press},
  year         = {2025}
}

@article{ref8,
  author       = {Jie Peng and
                  Yongxue Shan and
                  Yongfu Zha and
                  Xiaodong Wang},
  title        = {Adaptive multimodal graph learning for knowledge graph completion},
  journal      = {Data Min. Knowl. Discov.},
  volume       = {39},
  number       = {6},
  pages        = {70},
  year         = {2025},
  url          = {https://doi.org/10.1007/s10618-025-01139-x},
  doi          = {10.1007/S10618-025-01139-X},
  bibsource    = {dblp computer science bibliography, https://dblp.org}
}

@inproceedings{ref9,
  author       = {Yue Jian and
                  Xiangyu Luo and
                  Zhifei Li and
                  Miao Zhang and
                  Yan Zhang and
                  Kui Xiao and
                  Xiaoju Hou},
  editor       = {Toby Walsh and
                  Julie Shah and
                  Zico Kolter},
  title        = {{APKGC:} Noise-enhanced Multi-Modal Knowledge Graph Completion with
                  Attention Penalty},
  booktitle    = {Thirty-Ninth {AAAI} Conference on Artificial Intelligence, Thirty-Seventh
                  Conference on Innovative Applications of Artificial Intelligence,
                  Fifteenth Symposium on Educational Advances in Artificial Intelligence,
                  {AAAI} 2025, Philadelphia, PA, USA, February 25 - March 4, 2025},
  pages        = {15005--15013},
  publisher    = {{AAAI} Press},
  year         = {2025},
  url          = {https://doi.org/10.1609/aaai.v39i14.33645},
  doi          = {10.1609/AAAI.V39I14.33645},
  bibsource    = {dblp computer science bibliography, https://dblp.org}
}

@inproceedings{ref10,
    title = "Unleashing the Power of Imbalanced Modality Information for Multi-modal Knowledge Graph Completion",
    author = "Zhang, Yichi  and
      Chen, Zhuo  and
      Liang, Lei  and
      Chen, Huajun  and
      Zhang, Wen",
    editor = "Calzolari, Nicoletta  and
      Kan, Min-Yen  and
      Hoste, Veronique  and
      Lenci, Alessandro  and
      Sakti, Sakriani  and
      Xue, Nianwen",
    booktitle = "Proceedings of the 2024 Joint International Conference on Computational Linguistics, Language Resources and Evaluation (LREC-COLING 2024)",
    month = may,
    year = "2024",
    address = "Torino, Italia",
    publisher = "ELRA and ICCL",
    url = "https://aclanthology.org/2024.lrec-main.1487/",
    pages = "17120--17130",
}

@ARTICLE{ref11,
  author={Li, Linyu and Jin, Zhi and Zhang, Yichi and Jin, Dongming and Dou, Chengfeng and He, Yuanpeng and Zhang, Xuan and Zhao, Haiyan},
  journal={IEEE Transactions on Multimedia}, 
  title={Towards Structure-Aware Model for Multi-Modal Knowledge Graph Completion}, 
  year={2026},
  volume={28},
  number={},
  pages={1404-1416},
  doi={10.1109/TMM.2025.3639983}}

@inproceedings{ref12,
author = {Li, Xinhang and Zhao, Xiangyu and Xu, Jiaxing and Zhang, Yong and Xing, Chunxiao},
title = {IMF: Interactive Multimodal Fusion Model for Link Prediction},
year = {2023},
isbn = {9781450394161},
publisher = {Association for Computing Machinery},
address = {New York, NY, USA},
url = {https://doi.org/10.1145/3543507.3583554},
doi = {10.1145/3543507.3583554},
booktitle = {Proceedings of the ACM Web Conference 2023},
pages = {2572–2580},
numpages = {9},
location = {Austin, TX, USA},
series = {WWW '23}
}

@inproceedings{ref14,
author = {Zhang, Yichi and Chen, Zhuo and Guo, Lingbing and Xu, Yajing and Zhang, Wen and Chen, Huajun},
title = {Making Large Language Models Perform Better in Knowledge Graph Completion},
year = {2024},
isbn = {9798400706868},
publisher = {Association for Computing Machinery},
address = {New York, NY, USA},
url = {https://doi.org/10.1145/3664647.3681327},
doi = {10.1145/3664647.3681327},
booktitle = {Proceedings of the 32nd ACM International Conference on Multimedia},
pages = {233–242},
numpages = {10},
location = {Melbourne VIC, Australia},
series = {MM '24}
}

@article{ref13,
title = {GS-KGC: A generative subgraph-based framework for knowledge graph completion with large language models},
journal = {Information Fusion},
volume = {117},
pages = {102868},
year = {2025},
issn = {1566-2535},
doi = {https://doi.org/10.1016/j.inffus.2024.102868},
url = {https://www.sciencedirect.com/science/article/pii/S1566253524006468},
author = {Rui Yang and Jiahao Zhu and Jianping Man and Hongze Liu and Li Fang and Yi Zhou},
}

@InProceedings{ref16,
author="Liu, Yang
and Tian, Xiaobin
and Sun, Zequn
and Hu, Wei",
editor="Demartini, Gianluca
and Hose, Katja
and Acosta, Maribel
and Palmonari, Matteo
and Cheng, Gong
and Skaf-Molli, Hala
and Ferranti, Nicolas
and Hern{\'a}ndez, Daniel
and Hogan, Aidan",
title="Finetuning Generative Large Language Models with Discrimination Instructions for Knowledge Graph Completion",
booktitle="The Semantic Web -- ISWC 2024",
year="2025",
publisher="Springer Nature Switzerland",
address="Cham",
pages="199--217",
isbn="978-3-031-77844-5"
}

@inproceedings{ref15,
    title = "{KICGPT}: Large Language Model with Knowledge in Context for Knowledge Graph Completion",
    author = "Wei, Yanbin  and
      Huang, Qiushi  and
      Zhang, Yu  and
      Kwok, James",
    editor = "Bouamor, Houda  and
      Pino, Juan  and
      Bali, Kalika",
    booktitle = "Findings of the Association for Computational Linguistics: EMNLP 2023",
    month = dec,
    year = "2023",
    address = "Singapore",
    publisher = "Association for Computational Linguistics",
    url = "https://aclanthology.org/2023.findings-emnlp.580/",
    doi = "10.18653/v1/2023.findings-emnlp.580",
    pages = "8667--8683"
}

@inproceedings{ref17,
    title = "Enhancing Large Language Model for Knowledge Graph Completion via Structure-Aware Alignment-Tuning",
    author = "Liu, Yu  and
      Cao, Yanan  and
      Lin, Xixun  and
      Shang, Yanmin  and
      Wang, Shi  and
      Pan, Shirui",
    editor = "Christodoulopoulos, Christos  and
      Chakraborty, Tanmoy  and
      Rose, Carolyn  and
      Peng, Violet",
    booktitle = "Proceedings of the 2025 Conference on Empirical Methods in Natural Language Processing",
    month = nov,
    year = "2025",
    address = "Suzhou, China",
    publisher = "Association for Computational Linguistics",
    url = "https://aclanthology.org/2025.emnlp-main.1061/",
    doi = "10.18653/v1/2025.emnlp-main.1061",
    pages = "20970--20984",
    ISBN = "979-8-89176-332-6"
}

@inproceedings{ref18,
author = {Bordes, Antoine and Usunier, Nicolas and Garcia-Dur\'{a}n, Alberto and Weston, Jason and Yakhnenko, Oksana},
title = {Translating embeddings for modeling multi-relational data},
year = {2013},
publisher = {Curran Associates Inc.},
address = {Red Hook, NY, USA},
booktitle = {Proceedings of the 27th International Conference on Neural Information Processing Systems - Volume 2},
pages = {2787–2795},
numpages = {9},
location = {Lake Tahoe, Nevada},
series = {NIPS'13}
}

@inproceedings{ref19,
title     = {Embedding Entities and Relations for Learning and Inference in Knowledge Bases},
author    = {Yang, Bishan and Yih, Wen-tau and He, Xiaodong and Gao, Jianfeng and Deng, Li},
booktitle = {International Conference on Learning Representations},
year      = {2015}
}

@inproceedings{ref20,
title     = {Complex Embeddings for Simple Link Prediction},
author    = {Trouillon, Th{'e}o and Welbl, Johannes and Riedel, Sebastian and Gaussier, Eric and Bouchard, Guillaume},
booktitle = {Proceedings of the 33rd International Conference on Machine Learning},
series    = {Proceedings of Machine Learning Research},
volume    = {48},
pages     = {2071--2080},
year      = {2016},
publisher = {PMLR}
}

@inproceedings{ref21,
title     = {RotatE: Knowledge Graph Embedding by Relational Rotation in Complex Space},
author    = {Sun, Zhiqing and Deng, Zhi-Hong and Nie, Jian-Yun and Tang, Jian},
booktitle = {International Conference on Learning Representations},
year      = {2019}
}

@inproceedings{ref22,
    title = "{T}uck{ER}: Tensor Factorization for Knowledge Graph Completion",
    author = "Balazevic, Ivana  and
      Allen, Carl  and
      Hospedales, Timothy and abc",
    editor = "Inui, Kentaro  and
      Jiang, Jing  and
      Ng, Vincent  and
      Wan, Xiaojun",
    booktitle = "Proceedings of the 2019 Conference on Empirical Methods in Natural Language Processing and the 9th International Joint Conference on Natural Language Processing (EMNLP-IJCNLP)",
    month = nov,
    year = "2019",
    address = "Hong Kong, China",
    publisher = "Association for Computational Linguistics",
    url = "https://aclanthology.org/D19-1522/",
    doi = "10.18653/v1/D19-1522",
    pages = "5185--5194"
}

@inproceedings{ref23,
title     = {Image-embodied Knowledge Representation Learning},
author    = {Xie, Ruobing and Liu, Zhiyuan and Luan, Huanbo and Sun, Maosong},
booktitle = {Proceedings of the Twenty-Sixth International Joint Conference on Artificial Intelligence},
pages     = {3140--3146},
year      = {2017},
doi       = {10.24963/ijcai.2017/438}
}

@inproceedings{ref24,
title     = {Is Visual Context Really Helpful for Knowledge Graph? A Representation Learning Perspective},
author    = {Wang, Meng and Wang, Sen and Yang, Han and Zhang, Zheng and Chen, Xi and Qi, Guilin},
booktitle = {Proceedings of the 29th ACM International Conference on Multimedia},
pages     = {2735--2743},
year      = {2021},
publisher = {Association for Computing Machinery},
doi       = {10.1145/3474085.3475470}
}

@inproceedings{ref25,
title     = {{OTKGE}: Multi-modal Knowledge Graph Embeddings via Optimal Transport},
author    = {Cao, Zongsheng and Xu, Qianqian and Yang, Zhiyong and He, Yuan and Cao, Xiaochun and Huang, Qingming},
booktitle = {Advances in Neural Information Processing Systems},
volume    = {35},
pages     = {39090--39102},
year      = {2022}
}

@inproceedings{ref27,
title     = {{TIVA-KG}: A Multimodal Knowledge Graph with Text, Image, Video and Audio},
author    = {Wang, Xin and Meng, Benyuan and Chen, Hong and Meng, Yuan and Lv, Ke and Zhu, Wenwu},
booktitle = {Proceedings of the 31st ACM International Conference on Multimedia},
pages     = {2391--2399},
year      = {2023},
publisher = {Association for Computing Machinery},
doi       = {10.1145/3581783.3612266}
}

@inproceedings{ref28,
title     = {{VISTA}: Visual-Textual Knowledge Graph Representation Learning},
author    = {Lee, Jaejun and Chung, Chanyoung and Lee, Hochang and Jo, Sungho and Whang, Joyce Jiyoung},
booktitle = {Findings of the Association for Computational Linguistics: EMNLP 2023},
pages     = {7314--7328},
year      = {2023},
address   = {Singapore},
publisher = {Association for Computational Linguistics},
doi       = {10.18653/v1/2023.findings-emnlp.488}
}

@InProceedings{ref29,
  title = 	 {Knowledge Graph Large Language Model (KG-LLM) for Link Prediction},
  author =       {Shu, Dong and Chen, Tianle and Jin, Mingyu and Zhang, Chong and Du, Mengnan and Zhang, Yongfeng},
  booktitle = 	 {Proceedings of the 16th Asian Conference on Machine Learning},
  pages = 	 {143--158},
  year = 	 {2025},
  editor = 	 {Nguyen, Vu and Lin, Hsuan-Tien},
  volume = 	 {260},
  series = 	 {Proceedings of Machine Learning Research},
  month = 	 {05--08 Dec},
  publisher =    {PMLR},
  url = 	 {https://proceedings.mlr.press/v260/shu25a.html},
}

@inproceedings{ref30,
author = {Xie, Xin and Zhang, Ningyu and Li, Zhoubo and Deng, Shumin and Chen, Hui and Xiong, Feiyu and Chen, Mosha and Chen, Huajun},
title = {From Discrimination to Generation: Knowledge Graph Completion with Generative Transformer},
year = {2022},
isbn = {9781450391306},
publisher = {Association for Computing Machinery},
address = {New York, NY, USA},
url = {https://doi.org/10.1145/3487553.3524238},
doi = {10.1145/3487553.3524238},
booktitle = {Companion Proceedings of the Web Conference 2022},
pages = {162–165},
numpages = {4},
location = {Virtual Event, Lyon, France},
series = {WWW '22}
}

@inproceedings{ref31,
author = {Bordes, Antoine and Usunier, Nicolas and Garcia-Dur\'{a}n, Alberto and Weston, Jason and Yakhnenko, Oksana},
title = {Translating embeddings for modeling multi-relational data},
year = {2013},
publisher = {Curran Associates Inc.},
address = {Red Hook, NY, USA},
booktitle = {Proceedings of the 27th International Conference on Neural Information Processing Systems - Volume 2},
pages = {2787–2795},
numpages = {9},
location = {Lake Tahoe, Nevada},
series = {NIPS'13}
}

@inproceedings{ref32,
    title = "Unleashing the Power of Imbalanced Modality Information for Multi-modal Knowledge Graph Completion",
    author = "Zhang, Yichi  and
      Chen, Zhuo  and
      Liang, Lei  and
      Chen, Huajun  and
      Zhang, Wen",
    editor = "Calzolari, Nicoletta  and
      Kan, Min-Yen  and
      Hoste, Veronique  and
      Lenci, Alessandro  and
      Sakti, Sakriani  and
      Xue, Nianwen",
    booktitle = "Proceedings of the 2024 Joint International Conference on Computational Linguistics, Language Resources and Evaluation (LREC-COLING 2024)",
    month = may,
    year = "2024",
    address = "Torino, Italia",
    publisher = "ELRA and ICCL",
    url = "https://aclanthology.org/2024.lrec-main.1487/",
    pages = "17120--17130",
}

@inproceedings{ref33,
    title = "Noise-powered Multi-modal Knowledge Graph Representation Framework",
    author = "Chen, Zhuo  and
      Fang, Yin  and
      Zhang, Yichi  and
      Guo, Lingbing  and
      Chen, Jiaoyan  and
      Pan, Jeff Z.  and
      Chen, Huajun  and
      Zhang, Wen",
    editor = "Rambow, Owen  and
      Wanner, Leo  and
      Apidianaki, Marianna  and
      Al-Khalifa, Hend  and
      Eugenio, Barbara Di  and
      Schockaert, Steven",
    booktitle = "Proceedings of the 31st International Conference on Computational Linguistics",
    month = jan,
    year = "2025",
    address = "Abu Dhabi, UAE",
    publisher = "Association for Computational Linguistics",
    url = "https://aclanthology.org/2025.coling-main.11/",
    pages = "141--155"
}

@inproceedings{ref34,
author = {Hu, Edward J. and Shen, Yelong and Wallis, Phillip and Allen-Zhu, Zeyuan and Yuanzhi  Li and Wang, Shean and Wang, Lu and Chen, Weizhu},
title = {LoRA: Low-Rank Adaptation of Large Language Models},
booktitle = {ICLR 2022},
year = {2022},
month = {April},
url = {https://www.microsoft.com/en-us/research/publication/lora-low-rank-adaptation-of-large-language-models/},
}

@InProceedings{ref35,
author="Niu, Guanglin
and Zhang, Xiaowei",
editor="Zhu, Feida
and Yu, Philip S.
and Nadamoto, Akiyo
and Lim, Ee-Peng
and Shim, Kyuseok
and Ding, Wei
and Zhang, Bingxue",
title="Diffusion-Based Hierarchical Negative Sampling for Multimodal Knowledge Graph Completion",
booktitle="Database Systems for Advanced Applications",
year="2026",
publisher="Springer Nature Singapore",
address="Singapore",
pages="479--495",
isbn="978-981-95-3906-2"
}
\end{document}